\documentclass[10pt,twocolumn,letterpaper]{article}

\usepackage[pagenumbers]{cvpr}

\usepackage[dvipsnames]{xcolor}

\usepackage{color, colortbl}
\usepackage{xcolor}
\definecolor{Lightgray}{gray}{0.9}
\newcolumntype{a}{>{\columncolor{Lightgray}}c}

\newcommand{\pseudoparagraph}[1]{\vspace{-5mm}\noindent\paragraph{#1}}
\usepackage{makecell,booktabs,multirow}

\makeatletter
\newcommand{\largedagger}{\raisebox{.5ex}{\scriptsize\textdagger}}
\newcommand\notsotiny{\@setfontsize\notsotiny\@vipt\@viipt}
\makeatother

\usepackage{csquotes}

\newcommand{\bd}[1]{\textbf{#1}}
\newcommand{\ul}[1]{\underline{#1}}
\newcommand{\myddag}{\textsuperscript{\textdaggerdbl}}

\usepackage{tikz}
\usepackage{figs/aircraftshapes}
\usetikzlibrary{calc,positioning,3d,matrix,shadows,fit,backgrounds}
\usetikzlibrary{decorations.pathreplacing}
\usetikzlibrary{arrows.meta}
\usetikzlibrary{shapes.geometric}

\usepackage{float}

\usepackage{adjustbox}
\newcommand\mycolorbox[1]{\adjustbox{margin=0pt {\fboxsep},bgcolor=#1}}

\usepackage[pagebackref,breaklinks,colorlinks]{hyperref}

\title{Deep Event Visual Odometry}

\author{Simon Klenk\textsuperscript{1,2*} \quad Marvin Motzet\textsuperscript{1,2*} \quad Lukas Koestler\textsuperscript{1,2} \quad Daniel Cremers\textsuperscript{1,2}\smallskip\\
\textsuperscript{1}Technical University of Munich \quad \textsuperscript{2}Munich Center for Machine Learning\\
{\tt\small \{simon.klenk, marvin.motzet, lukas.koestler, cremers\}@tum.de}}

\begin{document}
\maketitle
\renewcommand{\thefootnote}{\fnsymbol{footnote}}
\footnotetext[1]{Equal contribution. Code at: \href{https://github.com/tum-vision/DEVO}{https://github.com/tum-vision/DEVO}}
\renewcommand{\thefootnote}{\arabic{footnote}}

\begin{abstract}
Event cameras offer the exciting possibility of tracking the camera's pose during high-speed motion and in adverse lighting conditions. Despite this promise, existing event-based monocular visual odometry (VO) approaches demonstrate limited performance on recent benchmarks. To address this limitation, some methods resort to additional sensors such as IMUs, stereo event cameras, or frame-based cameras.
Nonetheless, these additional sensors limit the application of event cameras in real-world devices since they increase cost and complicate system requirements. Moreover, relying on a frame-based camera makes the system susceptible to motion blur and HDR. To remove the dependency on additional sensors and to push the limits of using only a single event camera, we present Deep Event VO (DEVO), the first monocular event-only system with strong performance on a large number of real-world benchmarks. DEVO sparsely tracks selected event patches over time. A key component of DEVO is a novel deep patch selection mechanism tailored to event data. We significantly decrease the pose tracking error on seven real-world benchmarks by up to 97\% compared to event-only methods and often surpass or are close to stereo or inertial methods.
\end{abstract}    
\vspace{-4pt}
\section{Introduction}
\label{sec:intro}
Event cameras are visual sensors with high temporal resolution, high dynamic range, low latency, and low energy consumption. Hence, event cameras are ideally suited for unlocking several spatial computing applications in robotics or AR/VR, which are currently inaccessible for traditional frame-based cameras. A key component for spatial computing is visual odometry (VO), which estimates 6-dof camera poses from a visual data stream~\cite{cadena2016past}. The estimated poses are required by different downstream tasks, \eg, motion planning and control in robotics, or photorealistic rendering of virtual objects in VR applications. A primary challenge of existing VO systems is their limited robustness to motion blur and high dynamic range (HDR). These problematic scenarios can be substantially alleviated by utilizing an event camera instead of a conventional frame-based camera. Event cameras report per-pixel, asynchronous brightness changes above or below a camera-internal threshold with microsecond resolution, resulting in a large dynamic range. Therefore, event cameras are less susceptible to motion blur and HDR~\cite{gallego2020Survey, ultimateSlam18}.

Existing event-based VO methods~\cite{hidalgo22EDS, ultimateSlam18, guan22plevio, chen2023esvio, rebecq16EVO} have shown to increase the robustness of camera pose estimation in adverse conditions. However, most of these methods rely on additional sensors for good performance, such as a frame-based camera~\cite{hidalgo22EDS}, an IMU~\cite{guan22plevio}, an IMU and a frame-based camera~\cite{ultimateSlam18}, or an IMU with a stereo event camera setup~\cite{chen2023esvio}. Relying on additional sensors has the disadvantage of increased cost and complicated system requirements, \eg, requiring more space, higher energy density for powering all sensors, and more complicated calibration routines. Furthermore, systems relying on a frame-based camera are susceptible to motion blur and HDR.

In this work, we revisit the task of monocular event-only VO. We are motivated by the question:~\textit{What is the limit of general, real-world monocular event-only VO using no additional sensors?} Our work is inspired by the fact that events are close to optical flow in nature since events can be first-order approximated as dot product between optical flow and image brightness gradient~\cite{gallego2020Survey}. Hence, we propose to base our VO system on optical flow estimation, inspired by frame-based systems such as BASALT~\cite{usenko2019Basalt}, DROID-SLAM~\cite{teed2021droid}, DPVO~\cite{teed2022dpvo}, DF-VO~\cite{zhan2020WhatShouldBe}, and Tartan-VO~\cite{wang2021tartanvo}.

\noindent The main contributions of this paper are: 
\begin{itemize}
\item We introduce a novel patch selection mechanism specifically tailored towards event data, which increases the accuracy and robustness of DEVO.
\item DEVO is the first monocular, event-only method showing strong performance across \textit{seven} real-world benchmarks.
\item DEVO demonstrates that supervised learning on a large-scale dataset of simulated events enables strong generalization to real-world event VO benchmarks.
\item We open-source our code (including training, evaluation, and event data generation) to foster further research in deep event-based VO.
\end{itemize}


\section{Related Work} 
\label{sec:relatedWork}

\pseudoparagraph{Learning-based VO} 
Learning-based VO has shown to improve accuracy and robustness compared to classical, model-based VO~\cite{kazerouni2022survey}. Earlier approaches predict the pose end-to-end from images~\cite{wang2017deepvo, zhou2018deeptam, ummenhofer2017demon}. Drawing inspiration from the classical methods~\cite{mur2017orb2, engel2017direct}, a beneficial approach has emerged to first predict correspondences and subsequently use these in a principled, geometry-based manner for pose prediction~\cite{zhan2020WhatShouldBe, wang2021tartanvo, teed2021droid, teed2022dpvo}. This method utilizes the 3D geometry of the problem. Thus, it can avoid overfitting and increase robustness and generalization. Correspondence can either be explicitly modeled by feature points~\cite{mur2017orb2, forster2014svo, han2020superpointvo, detone2018self} or implicitly defined by optical flow~\cite{usenko2019Basalt, teed2021droid, teed2022dpvo, zhan2020WhatShouldBe, wang2021tartanvo}.

DROID-SLAM~\cite{teed2021droid} proposes to incorporate a recurrent update operator~\cite{teed2020raft} for iterative optical flow prediction and a differentiable bundle adjustment layer for pose estimation. DPVO~\cite{teed2022dpvo} proposes a sparse version for the DROID-SLAM frontend, tracking sparse patches randomly extracted from RGB frames and estimating optical flow for these patches only. Sparsification improves memory usage and runtime. While DPVO shows strong performance, its evaluation is limited to two real-world datasets.
 
\pseudoparagraph{Event-based VO using additional sensors}
The majority of event-based VO systems rely on additional sensors. Weikersdorfer et al.~\cite{weikersdorfer2014event} are the first to demonstrate event and depth VO on a small-scale custom dataset. Similarly, Zuo et al.~\cite{zuo2022devo} employ a depth and event sensor. Due to the scarcity of datasets containing both modalities, they only evaluate on MVSEC~\cite{zhu18mvsec} and on their custom dataset.

Hidalgo et al.~\cite{hidalgo22EDS} combine frames and events using the generative event model and photometric bundle adjustment inspired by DSO~\cite{engel2017direct}. Similarly, Kueng et al.~\cite{Mueggler16IROS} detect Harris corners~\cite{harris1988combined} in DAVIS~\cite{brandliDAVIS} frames and track those using events. A drawback of both methods is that specifically designed camera setups are required, either through dual pixel architecture~\cite{Mueggler16IROS} or through an optical beamsplitter with a large form factor with reduced incoming light~\cite{hidalgo22EDS}. Moreover, the problem of relying on frames is that the overall system is still susceptible to motion blur and HDR.

Inertial measurement units (IMUs) are either employed in the event-VO backend~\cite{mueggler2018continuous, zihao2017event, guan22monoEIO} or also for improved event feature tracking~\cite{rebecq2017real, le2020idol, yuan2016fast}. The problem of using an IMU is that the IMU bias parameter estimation is often difficult to achieve with sufficient accuracy, even with extensive parameter tuning~\cite{von2022dmvio, buchanan22deepIMU}.
Further sensor combinations such as event and frames, events and IMU~\cite{ultimateSlam18, mahlknecht2022exploring, guan22plevio}, or stereo event cameras and IMU~\cite{chen2023esvio} have been proposed. Using additional sensors tends to increase robustness since different modalities can complement each other.

\pseudoparagraph{Monocular event-only VO} 
Estimating 6-dof poses from only event data is very challenging. Thus, some approaches rely on a known photometric 3D map~\cite{gallego15event, bryner2019event} or restrict the motion type to rotation-only~\cite{kim2021real, reinbacher2017real, liu2020globally, liu2021spatiotemporal}, planar~\cite{wang2022visual, gallego15event}, or forward-facing motion \cite{zhu19neuroeVO}. Kim et al.~\cite{kim2016real} propose an approach for full 6-dof pose tracking, utilizing three Kalman filters for estimating pose, intensity gradient, and depth, respectively. They assume a known contrast threshold and show qualitative results. The monocular event-only VO proposed by Rebecq et al.~\cite{rebecq16EVO} is purely geometry-based, utilizing event-ray reprojection~\cite{rebecq2016emvs}. EVO can perform well on small-scale scenes, but it is very parameter-sensitive and requires bootstrapping, which is either achieved through a planar, fronto-parallel scene assumption or by using a frame-based camera.

Zhu et al.~\cite{zhu2019unsupervised} and Ye et al.~\cite{ye2018unsupervised} employ a CNN to predict pose, flow, and depth using unsupervised learning on MVSEC~\cite{zhu18mvsec}. Dense CNNs sacrifice event sparsity. Moreover, both methods fail to show generalization beyond the indoor flying scenes of MVSEC, which are also used for their training. In a similar fashion, Gelen et al.~\cite{gelen2023artificial} propose a SLAM system that is trained and evaluated on scenes from the same simulated CARLA environment~\cite{carla17}. They utilize three CNNs for end-to-end pose, depth, and loop-closure estimation, respectively, and collapse the events data to 2D histograms, sacrificing their temporal resolution. 

In contrast to the above deep event VO systems, our work is the only approach utilizing event sparsity. Moreover, we do not perform end-to-end pose prediction but utilize intermediate optical flow and differentiable bundle adjustment as in DPVO~\cite{teed2022dpvo}. We are the only approach training on a very large dataset and thus are able to demonstrate generalization from simulation to several real-world benchmarks.
\section{Deep Event Visual Odometry (DEVO)}
\definecolor{light-blue}{HTML}{47abd8}
\definecolor{green-light}{RGB}{229,235,178}
\definecolor{orange-light}{RGB}{255,214,165}
\definecolor{blue-light}{RGB}{193,239,255}
\colorlet{fill01}{white}
\def\myslant{-0.05}

\begin{figure*}[tb]
    \centering
    
    \begin{tikzpicture}[
		scale=1.0,
		font=\footnotesize,
		semithick,
		on grid,
            align=center,
		>={Stealth[length=5pt]},
		myshape/.style={
			 draw, fill=light-blue!55, rounded corners=2pt, drop shadow
		},
            loss/.style={
              draw, fill=red!30, rounded corners=2pt, minimum height=1.7em, 
            },
            flow/.style={
              draw, fill=light-blue!55, rounded corners=2pt, minimum height=1.7em,
              minimum width=1.75cm
            },
		voxels/.pic = {
			\foreach \x in {1,...,5}
			\node[inner sep=0pt, outer sep=0pt, canvas is xy plane at z=0.2*\x-0.5, draw=black, yslant=\myslant] (-voxel-\x) {\includegraphics[trim={0 0.9cm 0 0.9cm},clip,width=2.5cm]{figs/imgs/evs0\x.png}};
			\node[anchor=south, label={below,yslant=\myslant}:{Event Voxel Grids $\mathbf{E}$}] at (-voxel-5.south) {};
		},
		patch/.pic = {
			\foreach \x in {1,...,5}
			\node[inner sep=0pt, outer sep=0pt, canvas is xy plane at z=0.2*\x-0.5, draw=black, yslant=\myslant] (-voxel-\x) {\includegraphics[trim={0 0.9cm 0 0.9cm},clip,width=2.5cm]{figs/imgs/patch0\x.png}};
			\node[anchor=south, label={below,yslant=\myslant}:{\textcolor{orange}{Patches} $\mathbf{P}_t$ on $\mathbf{E}_t$}] at (-voxel-5.south) {};
		},
		depth/.pic = {
			\node[inner sep=0pt, outer sep=0pt, canvas is xy plane at z=0.1, draw=black, yslant=\myslant
			] (-img) {\includegraphics[trim={0 0.9cm 0 0.9cm},clip,width=2.5cm]{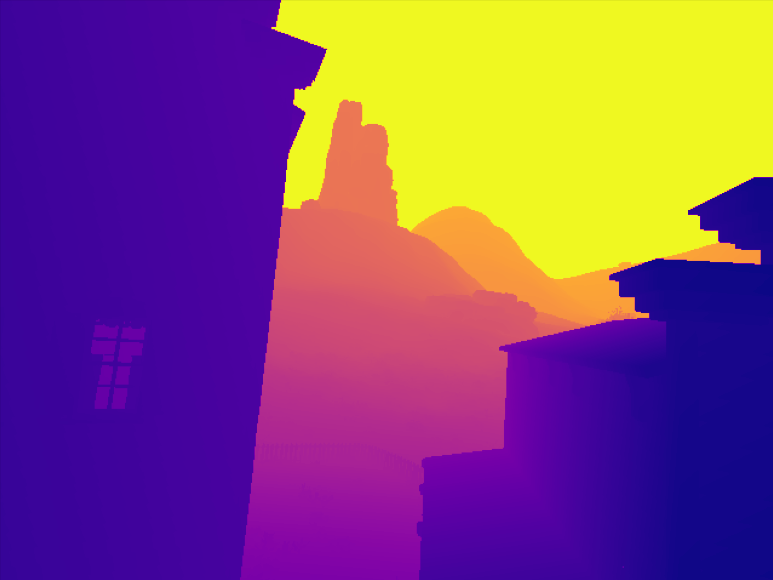}};
			\node[anchor=south, label={below,yslant=\myslant}:{Inverse Depths $\mathbf{d}$}] at (-img.south) {};
		},
		scorer/.pic = {
			\node[inner sep=0pt, outer sep=0pt, canvas is xy plane at z=0.1, draw=black, yslant=\myslant
			] (-img) {\includegraphics[trim={0 0.9cm 0 0.9cm},clip,width=2.5cm]{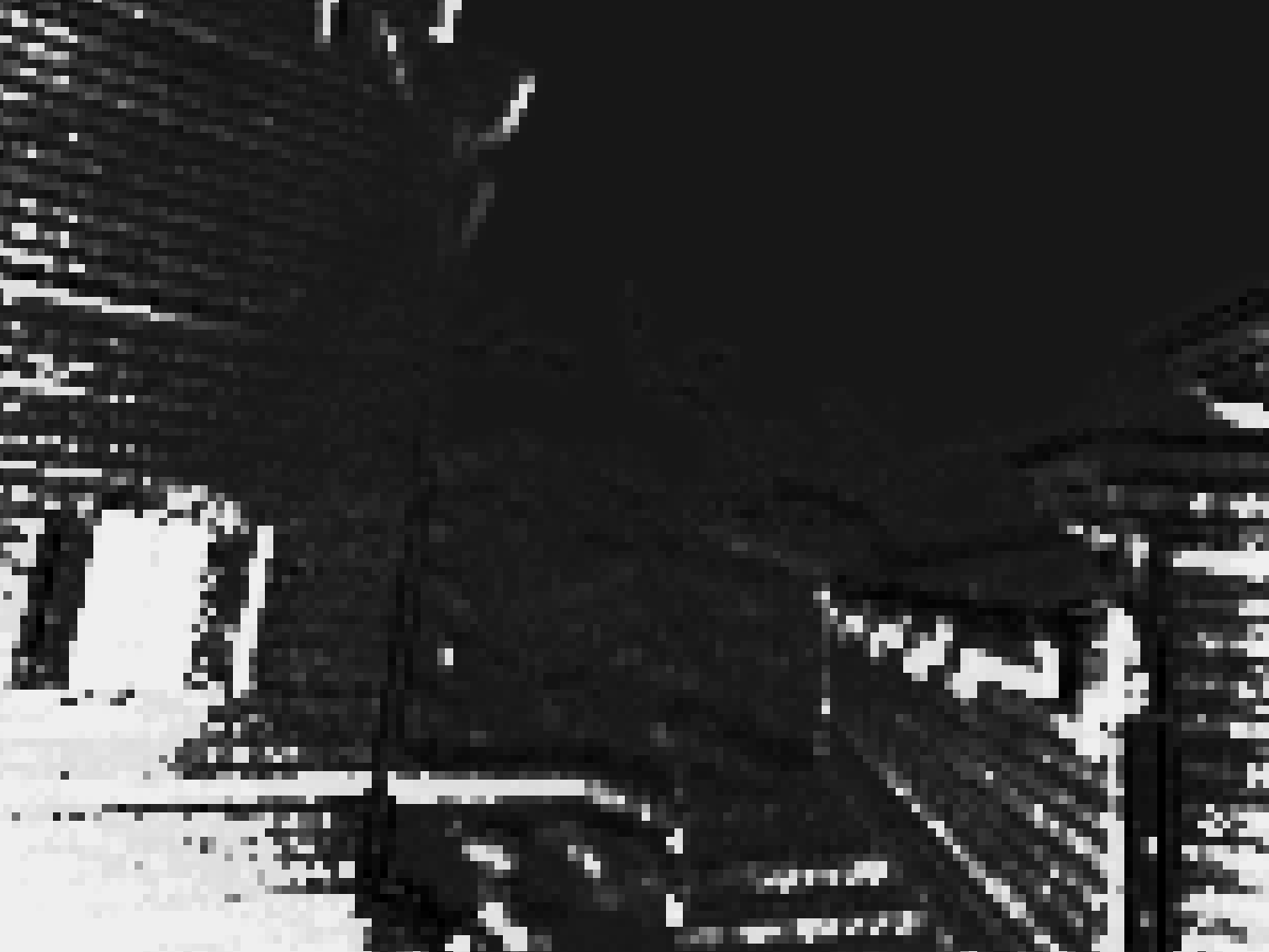}};
			\node[anchor=south, label={below,yslant=\myslant}:{Score Map $\mathbf{S}_t$}] at (-img.south) {};
	},
		pics/net/.style args={#1/#2/#3}{code = {
			\draw [myshape, inner sep=0pt, outer sep=0pt, shift={(-#1/2,0)}] (0,0) -- (0,#2) -- (#1,#3) -- (#1,-#3) -- (0,-#2) -- cycle;
			\coordinate (-center) at (0,0);
			\coordinate (-out) at (#1/2,0);
			\coordinate (-in) at (-#1/2,0);
			\node[] (-txt) at (0,0) {\bfseries Patch Selector};
		}},
		box/.pic = {
			\node[rectangle, inner sep=0pt, outer sep=0pt, myshape, minimum height=2cm, minimum width=2.5cm] (-box) at (0,0) {\bfseries Update Operator};
		},
		frust/.pic = {
                \coordinate (-center) at (0,0);
                \begin{scope}[shift={(-.75,0)},inner sep=0pt, outer sep=0pt]
                \node [quadcopter top,fill=white,draw=black,minimum width=.7cm,yslant=-.2] (-quad) at (0,0) {};
			\draw[densely dashed] (0.5,0.25) rectangle (1,0.75);
			\draw[densely dashed] (0,0) -- (1,.75);
			\draw[densely dashed] (0,0) -- (1,.25);
			\draw[densely dashed] (0,0) -- (.5,.75);
			\draw[densely dashed] (0,0) -- (.5,.25);
                \end{scope}
			\node[right=.8cm of -center,anchor=center,yslant=\myslant] (-txt) {Poses $\mathbf{T}$};
			\coordinate (-north) at (0,.75);
		},
		every shadow/.style={opacity=.2, shadow xshift=.5ex, shadow yshift=.5ex},
		]
		
		\matrix (m) [matrix of nodes, row sep=0.1cm, column sep=1cm,
		anchor=center,
		nodes={anchor=center, align=center},
		nodes in empty cells,
		row 4/.append style={nodes={text width=1.75cm, text depth=0ex,text height=1.5ex}},
		]{
			\draw pic[] (cam) {frust}; &
			& & & \node[myshape, minimum height=1.15cm, minimum width=2.5cm] (dba) {\bfseries DBA}; \\
			\draw pic (depth) {depth}; & & & & 
                \\
			\draw pic[] (vox) {voxels}; &
			\draw pic[] (ps) {net=2.0/0.9/.4}; &
			\draw pic (scorer-map) {scorer}; &
			\draw pic[] (patches) {patch}; & 
			\draw pic[] (devo) {box}; \\
			\textbf{Training Data} & & & &
			|[xshift=0em]|\textbf{Estimation} \\
		};

		\node[anchor=north] at (devo-box.north) (t-est) {$\big(\Delta\hat{\mathbf{f}}^{(i)},\omega^{(i)}\big)$};
		\coordinate (dpvo-aux) at ([xshift=1em]devo-box.east);
		\draw[->, rounded corners] ([xshift=-1em]devo-box.south east) |- ++(2em,-1em) -- ([xshift=1em,yshift=1em+.5ex]devo-box.north east) -| ++(-2em,-1em) {};
            \node[anchor=north] at ([yshift=-1em]devo-box.south east) {$\times 12$};
		
		\node[anchor=west] at (dba.west) {$\hat{\mathbf{T}}$\\$\hat{\mathbf{d}}$};
            \coordinate (dba-aux) at (dba.north|-cam-north);
		
		\draw[->, bend left=20] (scorer-map-img.north) to node[midway, above] (sampler) {Pooled Multinomial Sampling} (patches-voxel-5.north);
		
		\node[anchor=center, align=center, fit=(m-4-2)(m-4-3)(m-4-4), inner sep=0pt, outer sep=0pt, text depth=0ex,text height=1.5ex] (txt-patch) {\bfseries Patch Selector \& Sampler};

		\begin{scope}[on background layer, rounded corners=5pt, thick]
			\node[draw=green-light!300, fill=green-light!30, fit=(cam-north) (depth-img) (vox-voxel-1) (vox-voxel-5) (m-4-1)] (input) {};
			\node[draw=orange-light!300, fill=orange-light!30, fit=(ps-in) (patches-voxel-1) (patches-voxel-5) (m-4-2) (m-4-3) (txt-patch) (sampler)] (patch-selector) {};
			\node[draw=blue-light!300, fill=blue-light!30, fit=(devo-box) (t-est) (m-4-5) (dpvo-aux) (dba-aux) ] (output) {};
		\end{scope}

		\begin{scope}[rounded corners]
			\draw[<-] (ps-in) to node[above, midway] {$\mathbf{E}_t$} (ps-in-|vox-voxel-1.east);

			\draw[->] (ps-out) to node[above, midway] {$\mathbf{S}_t$} (ps-out-|scorer-map-img.west);

                \draw[<-] (devo-box.west) -- (devo-box.west-|patches-voxel-1.east)
			node[above, midway] {$\mathbf{E}_t$}
			node[below, midway] {$\mathbf{P}_t$};

			\draw[->] ([yshift=.5ex]devo-box.north) -- (dba.south);
			
                \node[loss] (pose-loss) at ($(dba.center-|cam-center)!0.5!(dba.center)$) {$\mathcal{L}_\textrm{pose}(\mathbf{T}, \hat{\mathbf{T}})$};

                \draw[->] (dba.west) -- (pose-loss.east);
                \draw[->] (input.east|-cam-txt.east) -- (pose-loss.west);
   
                \node[loss,anchor=north] (flow-loss) at ([yshift=-2ex]pose-loss.south) {$\mathcal{L}_\textrm{flow}(\mathbf{f},\hat{\mathbf{f}})$};

                \coordinate (mid) at (flow-loss-|input.east);
                \draw[decorate,decoration={brace,amplitude=5pt,raise=0pt,aspect=.5},xshift=0pt,yshift=0pt,rounded corners=.5pt] ([yshift=.75cm]mid) -- ([yshift=-.75cm]mid);

                \draw[->] ([xshift=5pt]mid) -- (mid-|patch-selector.west);
                \node[flow, anchor=west] (flow-gt) at (mid-|patch-selector.west) {$f(\mathbf{T},\mathbf{d})$};
                \node[anchor=south, inner sep=2pt, text centered] at (flow-gt.north) {Compute flow $\mathbf{f}$};

                \node[flow, anchor=east] (flow-est) at (patch-selector.east|-flow-loss.east) {$f(\hat{\mathbf{T}},\hat{\mathbf{d}})$};
                \node[anchor=south, inner sep=2pt] at (flow-est.north) {Compute flow $\hat{\mathbf{f}}$};

                \draw[->] (flow-gt.east) -- (flow-loss.west);
                \draw[->] (flow-est.west) -- (flow-loss.east);
                \draw[->] (dba.west) -| ($(dba.west)!0.5!(flow-est.east)$) |- (flow-est.east);
   
			\node[loss,anchor=north] (scorer-loss) at ([yshift=-2ex]flow-loss.south){$\mathcal{L}_\textrm{score}(\mathbf{f},\hat{\mathbf{f}},\mathbf{S},\omega)$};
   
                \draw[->] (flow-gt.south) |- (scorer-loss.west);
                \draw[->] (flow-est.south) |- (scorer-loss.east);
                \draw[->] (devo-box.north) |- (scorer-loss.east);
		\end{scope}	
	\end{tikzpicture}
    
    \caption{Overview of our proposed method. During training, DEVO takes event voxel grids $\{\mathbf{E}_t\}_{t=1}^N$, inverse depths $\{\mathbf{d}_t\}_{t=1}^N$, and camera poses $\{\mathbf{T}_t\}_{t=1}^N$ of a sequence of size $N$ as input. DEVO estimates poses $\{\hat{\mathbf{T}}_t\}_{t=1}^N$ and depths $\{\hat{\mathbf{d}}_t\}_{t=1}^N$ of the sequence. Our novel patch selection network predicts a score map $\mathbf{S}_t$ to highlight optimal 2D coordinates $\mathbf{P}_t$ for optical flow and pose estimation.
    A recurrent update operator iteratively refines the sparse patch-based optical flow $\hat{\mathbf{f}}$ between event grids by predicting $\Delta\hat{\mathbf{f}}$ and updates poses and depths through a differentiable bundle adjustment (DBA) layer, weighted by $\omega$, for each revision.
    Ground truth optical flow $\mathbf{f}$ for supervision is computed using poses and depth maps.
    At inference, DEVO samples from a multinomial distribution based on the pooled score map $\mathbf{S}_t$.}
    \vspace{-4pt}
    \label{fig:DEVO_figure}
\end{figure*}

Our approach extends DPVO~\cite{teed2022dpvo} to the event modality.
We propose a novel patch selection mechanism for sparse event data. Event data pose unique challenges, \eg, a large sim-to-real gap~\cite{stoffregen20sim2real}. We thus propose specific event augmentations during training and randomized event simulation.

\subsection{Event Representation}
Events are a stream of tuples $(x_k, y_k, t_k, p_k)$, indicating an increase ($p_k\! =\! 1$) or decrease ($p_k\! =\! -1$) of observed brightness at pixel $\mathbf{u} = (x_k, y_k)$ with microsecond time\-stamp $t_k$. We process events to a sequence of volumetric voxel grids $\{\mathbf{E}_t\}_{t=1}^N$~\cite{zhu2019unsupervised} for compatibility with standard neural networks, where $\mathbf{E}_t\in\mathbb{R}^{H\times W\times 5}$. Voxel grids preserve temporal information by bilinear interpolation of event counts in time. We discretize the time dimension into five bins and normalize each voxel grid to zero mean and unit variance.
Each event voxel grid $\mathbf{E}_t$ is assigned to a ground truth camera pose $\mathbf{T}_t\in\mathbb{SE}(3)$ and a ground truth inverse depth map $\mathbf{d}_t\in\mathbb{R}_+^{H\times W}$.
Note that in the following, $\hat{\mathbf{T}}_t$ and $\hat{\mathbf{d}}_t$ denote the predictions of $\mathbf{T}_t$ and $\mathbf{d}_t$, respectively.

\pseudoparagraph{Photometric voxel augmentations} 
The ESIM~\cite{rebecq2018esim} event simulator employs the event generation model~\cite{gallego2020Survey}
\begin{equation}\label{eq:evModel}
\Delta L(\mathbf{u}_k, t_k) = L(\mathbf{u}_k, t_k) - L(\mathbf{u}_k, t_{k-1}) = p_k C,
\end{equation} 
\noindent where $C$ is the camera-internal contrast threshold and $L(\mathbf{u}, t)\! =\! \ln(\mathbf{I}_t[\mathbf{u}])$ is the logarithmic mapping of image brightness. ESIM tends to produce densely populated event voxels, especially in dark image regions, due to the logarithm in~\cref{eq:evModel}. Hence, the training dataset contains noticeably more densely populated event voxels than the real-world evaluation datasets. To reduce the sim-to-real gap, we augment the simulated voxels used for training with photometric augmentations, which reduce event density.

\subsection{Deep Event Patch Selection}\label{sec:method}
Frame-based methods such as DSO~\cite{engel2017direct}, LSD-SLAM~\cite{engel2014lsd}, and DPVO~\cite{teed2022dpvo} successfully utilize patches covering most of the image plane, which works well for dense data like RGB images. However, events frequently only sparsely cover the image plane (\cf \cref{fig:DEVO_figure}), leaving large areas devoid of events. Tracking features in such non-discriminative areas often results in VO failures~\cite{gao2018ldso}.

\pseudoparagraph{Approach overview}
We re-use the patch graph architecture and the iterative update operator from DPVO~\cite{teed2022dpvo}. The dynamic patch graph $(\mathcal{V},\mathcal{E})$ connects event patches on $\mathbf{E}_t$ with event voxel grids $\mathbf{E}_{t'}$ such that $t'\neq t$. Patch trajectories, which form the correspondences for VO, are obtained by reprojecting a patch in all its connected event grids of the patch graph. The update operator, a recurrent neural network, iteratively proposes revisions $\Delta\hat{\mathbf{f}}$ to optical flow estimates $\hat{\mathbf{f}}$. Based on the most recent optical flow estimate, the differentiable bundle adjustment (DBA) layer~\cite{teed2022dpvo, tang2018ba, lindenberger2021pixel} updates camera poses and patch depths of all patches within a sliding window of recent keyframes.

\pseudoparagraph{Patch selection network}
We propose a novel \emph{patch selection network} to make sparse patch-based tracking accurate and robust for the event modality. The patch selection network predicts a \emph{score map} $\mathbf{S}_t\in[0,1]^{H/4\times W/4}$ of voxel grid $\mathbf{E}_t$, highlighting 2D coordinates which are optimal for optical flow and pose estimation. It consists of three convolutional layers of kernel size three, each followed by a ReLU. The voxel grids provide five input channels, which are progressively increased to $[8, 16, 32]$ channels, respectively. The fourth convolutional layer outputs one channel and is followed by max-pooling with kernel size and stride of four. The network outputs the one-channel score map bounded to $[0,1]$ by using a sigmoid activation.
The patch selection network is trained jointly with the overall system using a dedicated score loss $\mathcal{L}_\text{score}$ (\cf\cref{eq:scorer_loss}).

\pseudoparagraph{Training the score map}
We select $P$ patch coordinates $\mathbf{P}_t = \{\mathbf{p}_k^t\}_{k=1}^P$ per voxel grid $\mathbf{E}_t$ based on the score values $s_k = \mathbf{S}_t[\mathbf{p}_k^t]$, where $\mathbf{p}_k^t\in\mathbb{R}^{H/4 \times W/4}$.
We aim to predict high scores $s_k$ at patch coordinates $\mathbf{p}_k^t$ which are well-suited for tracking, \ie, the optical flow residuals $r_{kj}$ of patch $k$ onto voxel grid $\mathbf{E}_j$ are small \emph{and} the estimated confidence weights $\omega_{kj}$ of the DBA are large. To achieve this, we minimize score map values with large tracking error by multiplying the score $s_k$ of patch $k$ with the corresponding residuals $r_{kj}$ and \enquote{inverted weight} $(1 -\alpha\ln\omega_{kj})$. Large score map values are enforced by subtracting the logarithm of the sampled score map values $\mathbf{S}_{\mathbf{P}} = \mathbf{S}[\mathbf{P}]$:
\begin{equation}\label{eq:scorer_loss}
\mathcal{L}_{\text{score}} = \frac{1}{\lvert\mathcal{E}\rvert}\sum_{(k,j)\in\mathcal{E}} s_k r_{kj} (1 -\alpha\ln\omega_{kj}) - \ln\mathbf{S}_{\mathbf{P}}.
\end{equation}
\noindent The term $\alpha$ controls the influence of the DBA weights. 
Note that no ground truth labels for $\mathbf{S}$ are needed for training.

\begin{figure}
     \centering
   
     \begin{subfigure}[b]{0.155\textwidth}
          \centering
          \includegraphics[width=\textwidth]{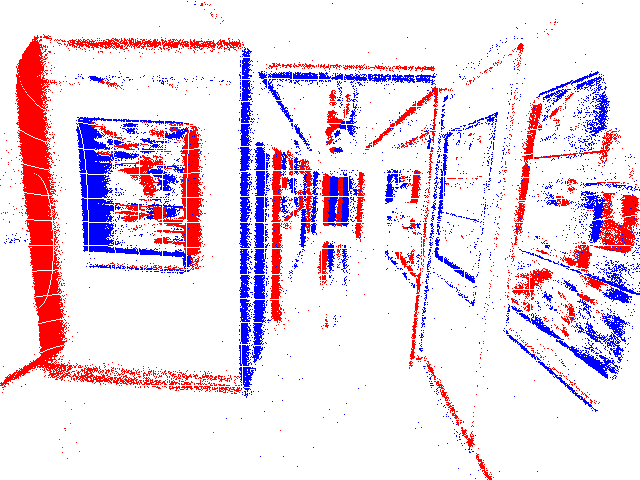}
          \captionsetup{font=footnotesize}
          \caption{Raw events}
      \end{subfigure}
     \begin{subfigure}[b]{0.155\textwidth}
          \centering
          \includegraphics[width=\textwidth]{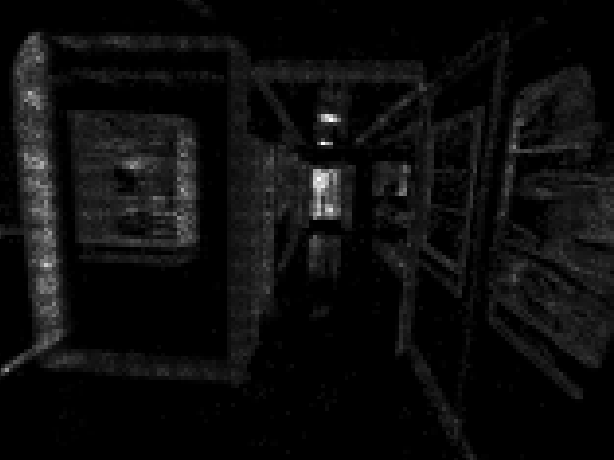}
          \captionsetup{font=footnotesize}
          \caption{Image gradient}
      \end{subfigure}
     \begin{subfigure}[b]{0.155\textwidth}
          \centering
          \includegraphics[width=\textwidth]{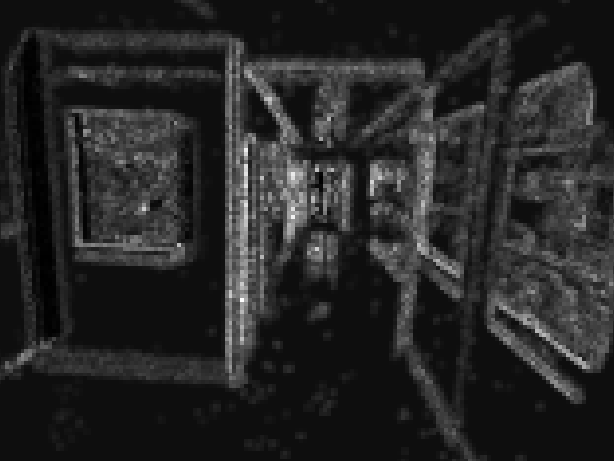}
          \captionsetup{font=footnotesize}
          \caption{Score map}
      \end{subfigure}
      
     \captionsetup{font=footnotesize}
     \caption{Comparison of gradient map (middle) and proposed score map (right) for event input (left). Selecting patches from our learned score map is more accurate and robust than from the gradient map (\cf~\cref{sec:ablation}).}
     \vspace{-12pt}
     \label{fig::scorer-map}
\end{figure}

During training, we first randomly evaluate $3 P$ patch coordinates and subsequently use the $P$ patches with the highest value in the score map for tracking. We refer to this strategy as \textit{$3P$-random}.
\Cref{fig::scorer-map} shows the qualitative difference between our learned score map and a gradient map.

The total loss is given by $\mathcal{L} = 0.05\mathcal{L}_\text{score} + 0.1\mathcal{L}_\text{flow} + 10\mathcal{L}_\text{pose}$.
While $\mathcal{L}_\text{flow}$ and $\mathcal{L}_\text{pose}$ are applied after each iteration of the update operator, $\mathcal{L}_\text{score}$ is only applied after the last iteration.
Ground truth optical flow of event patches is computed from camera poses and depth maps, see~\cref{fig:DEVO_figure}.

\pseudoparagraph{Sampling from the score map}
Sampled patches need to be broadly distributed over the entire image plane in order to achieve accurate VO~\cite{lu2017large, cai2018monocular, usenko2019Basalt}. Hence, we first subdivide the image plane into $G$ disjoint grid cells before sampling. Within each grid cell, we sample $\frac{P}{G}$ patches.

Moreover, we want to sample patches with high values in the score map since high values indicate relevant coordinates for tracking. This approach could be achieved by following the \emph{$3P$-random} strategy of the training or by simply picking the $P$ highest values in the score map (\emph{top-$P$ sampling}). 
However, both strategies are still sensitive to outliers, less accurate, and less robust on average (\cf~\cref{sec:ablation}) than our proposed \emph{pooled multinomial sampling}.

Multinomial sampling models a multinomial distribution of score map values over 2D patch coordinates and subsequently samples $P$ distinct 2D coordinates without replacement ($\frac{P}{G}$ per grid cell). Multinomial sampling likely chooses high values in the score map, but it does not only choose the highest values. Hence, it is less sensitive to real-world outliers of the score map than top-$P$ sampling.

Note that the score map's resolution $(H_{\text{sc}}, W_{\text{sc}})$ is one $1/4$-th of the input resolution $(H, W) = (4 H_{\text{sc}}, 4 W_{\text{sc}})$ as the feature map's resolution in DPVO~\cite{teed2022dpvo}.
To further robustify the sampling procedure, we propose \emph{pooled sampling}, which performs a $4\times 4$ average pooling with stride of four on the score map before multinomial sampling, resulting in a \emph{pooled score map} of $(H_{\text{pool}},  W_{\text{pool}}) = (H/16, W/16)$. This smoothens score map values and increases the receptive field. For each sampled value in the pooled score map, we apply a second multinomial sampling on the corresponding $4\times 4$ window of the initial score map to get 2D coordinates of the resolution $(H_{\text{sc}}, W_{\text{sc}})$.
Furthermore, pooled sampling has a performance advantage as sampling is performed at a small resolution $(H_{\text{pool}},  W_{\text{pool}})$. Since we sample distinct 2D coordinates  without replacement in the pooled score map and patches in the score map are of size $3 \times 3$, we automatically avoid overlapping patches in the score map. The sampled patch coordinates are used to index the matching and context feature map as in DPVO~\cite{teed2022dpvo}.

\pseudoparagraph{Implementation details}
We simulate events on all sequences of the TartanAir dataset~\cite{tartanair20iros} using the ESIM~\cite{rebecq2018esim} simulator. We randomize the contrast thresholds per sequence, sampling the negative and positive contrast thresholds independently from a uniform distribution in $C \sim \mathcal{U}(0.16, 0.34)$. We perform multi-GPU training for a total of 240,000 iterations on two A40s with a batch size of 1, sequence size $N = 15$, and $P = 80$, which takes 2.5 days.
We apply a $2\times 2$ grid for sampling during evaluation.
\section{Experiments}
We evaluate DEVO on seven real-world event VO benchmarks while training in simulation only.
We compare our method to multiple baselines from the literature, including methods using more sensors. We propose two additional baselines: (i)~DPVO~\cite{teed2022dpvo}, which runs on the RGB frames of the respective dataset, (ii)~DPVO\largedagger~\cite{rebecq2019e2vid, teed2022dpvo}, which is a re-trained DPVO model on the E2VID~\cite{rebecq2019e2vid} reconstructions of all TartanAir sequences. The re-trained model runs on E2VID reconstructions of the respective target dataset. Different from DPVO~\cite{teed2022dpvo}, DEVO and the baselines (i) and (ii) are trained on \textit{all} sequences of the TartainAir~\cite{tartanair20iros} dataset. 
To obtain high-quality reconstructions for DPVO\largedagger~\cite{rebecq2019e2vid, teed2022dpvo}, we only feed a subset of all events per grid into E2VID~\cite{rebecq2019e2vid} for fast or high-resolution sequences.

\subsection{Quantitative Evaluation}
We intentionally refrain from parameter tuning to demonstrate generalization. We thus use the exact same parameter setting for all evaluation sequences, except the keyframe threshold, which we set to 5 on RPG~\cite{zhou18RPG}, MVSEC~\cite{zhu18mvsec} and UZH-FPV~\cite{delmer19FPV}, to 25 on EDS~\cite{hidalgo22EDS} and to 15 on all other datasets. The keyframe threshold is specified in terms of the average optical flow, which is smaller for smaller resolution and slower motions.

\begin{table*}[tb]
\centering
\small

\begin{tabular}{l l c c c c c c c c c c}
\toprule
& \multicolumn{1}{c}{} & \multicolumn{6}{c}{indoor\_fwd} & \multicolumn{3}{c}{indoor\_45\_deg} \\
\cmidrule(r){3-8} \cmidrule(l){9-11}

Method & Modality &  3 & 5 & 6 & 7 & 9 & 10 & 2 & 4 & 9 \\
\midrule

\rowcolor{Lightgray} 
ORB-SLAM3~\cite{campos21orb3} & Stereo VIO  & 0.55 & 1.19 & -- & \bd{0.36} & 0.77 & 1.02 & 2.18 & 1.53 & \bd{0.49} \\
\rowcolor{Lightgray} 
VINS Fusion~\cite{qin19VINS} & Stereo VIO & 0.84 & -- & 1.45 & 0.61 & 2.87 & 4.48 & -- & -- & -- \\
\rowcolor{Lightgray} 
VINS Mono~\cite{qin18VINSMONO}& Mono VIO & 0.65 & 1.07 & \bd{0.25} & \ul{0.37} & 0.51 & 0.92 & \bd{0.53} & 1.72 & 1.25 \\
\rowcolor{Lightgray} 
DPVO~\cite{teed2022dpvo} & Mono VO & -- & -- & -- & -- & -- & -- & -- & -- & -- \\

\midrule

\rowcolor{Lightgray} 
USLAM~\cite{ultimateSlam18} & Mono EVIO & -- & -- & -- & -- & -- & -- & -- & 9.79 & 4.74 \\
\rowcolor{Lightgray} 
PLEVIO~\cite{guan22plevio} & Mono EVIO & \ul{0.38} & 0.90 & \ul{0.30} & 0.55 & \bd{0.44} & 1.06 & \ul{0.55} & 1.30 & \ul{0.76} \\
EVO~\cite{rebecq16EVO} &  Mono EO & -- & -- & -- & -- & -- & -- & -- & -- & -- \\
DPVO\largedagger &  Mono EO\largedagger & 0.52 & \ul{0.42} & 0.55 & -- & \ul{0.45} & \ul{0.54}  & -- & \ul{1.21} & -- \\
\textbf{DEVO (ours)} & Mono EO & \bd{0.37} & \bd{0.40} & 0.31 & 0.50 & 0.61 & \bd{0.52} & 0.72 & \bf{0.45} & 0.89 \\
\bottomrule
\end{tabular}
\caption{UZH-FPV drone racing dataset~\cite{delmer19FPV} with MPE[\%/m]. Baseline numbers are taken from~\cite{guan22plevio}.}
\label{tab::FPV}
\vspace{-12pt}
\end{table*}

On each dataset, we run five trials and report the median result. For monocular methods, the global scale is estimated once before alignment. We employ absolute trajectory error ATE[cm] ($\downarrow$)~\cite{sturm12benchmark}, RMSE rotational error $\mathbf{R}_{\text{rmse}}[\text{deg}]$ ($\downarrow$)~\cite{zhang18tutorial}, and mean position error MPE[\%/m] ($\downarrow$)~\cite{ultimateSlam18, chen2023esvio}, utilizing the EVO toolbox~\cite{grupp2017evo}. For stereo datasets, we employ the left camera stream.
In the following tables, we highlight methods with access to additional sensors in \mycolorbox{Lightgray}{gray} and mark results with \enquote{--} to indicate a failure.

\pseudoparagraph{UZH-FPV drone racing dataset}
The UZH-FPV drone racing dataset~\cite{delmer19FPV} provides data from a DAVIS346~\cite{brandliDAVIS} mounted on a drone performing aggressive maneuvers at high speed.
\Cref{tab::FPV} shows that our method outperforms all related work on four of nine sequences. This is noteworthy because all other successful methods use an IMU, except our proposed event-only baseline DPVO\largedagger.
DPVO\largedagger~fails on three of nine sequences and performs worse than DEVO on eight of nine sequences. VINS Fusion~\cite{qin19VINS} and ORB-SLAM3~\cite{campos21orb3} both employ stereo VIO, but usually they perform worse than DEVO. The event-based methods USLAM~\cite{ultimateSlam18} and PLEVIO~\cite{guan22plevio} have access to strictly more data than our method. However, both methods are still outperformed by DEVO on the majority of the sequences. We compare DEVO to other monocular methods, DPVO~\cite{teed2022dpvo} and EVO~\cite{rebecq16EVO}, which both fail on all sequences.

\pseudoparagraph{VECtor}
\bgroup

\begin{table*}[tb]
\centering
\scriptsize

\begin{tabular}{p{1.5cm} aaa | aaa ccc}
\toprule
Method & ORB3~\cite{campos21orb3} & VINSFusion~\cite{qin19VINS} & DPVO~\cite{teed2022dpvo} & ESVIO~\cite{chen2023esvio} & PLEVIO~\cite{guan22plevio} & ESVO~\cite{zhou21ESVO} & EVO~\cite{rebecq16EVO} & DPVO\textsuperscript{\textdagger} & \bd{DEVO(ours)} \\

Modality & StereoVIO & StereoVIO & MonoVO & StereoEVIO & MonoEVIO & StereoEO & MonoEO & MonoEO\largedagger &  MonoEO \\

\cmidrule(){2-10}
& MPE/ATE & MPE/ATE & MPE/ATE & MPE/ATE & MPE/ATE & MPE/ATE & MPE/ATE & MPE/ATE & MPE/ATE 
\\

\midrule

corner-slow    &   1.49 / \ul{1.2}   &   1.61 / 1.3    &   \bd{0.30} / \bd{0.4}   &   1.49  / \ul{1.2}    &    2.10 / 1.7    &     4.83 / 3.9   &    4.33 / 3.5   &  -- / --  &  \ul{0.59} / \ul{1.2} \\

robot-norm   &   0.73 / 2.9   &    0.58 / 2.3   &    \bd{0.15} / \bd{0.7}    &    1.08 / 4.3     &    0.68 / 2.7   &   -- / --   &   3.25 / 13.0  &  0.22 / 2.4 &  \ul{0.17} / \ul{1.0} \\
robot-fast     &   0.71 / 15.0   &    -- / --      &  \bd{0.07} / \bd{1.7}     &  0.20 / 4.2 &  0.17  / \ul{3.7} &  -- / --  & -- / --  & 0.73 / 18.9 & \ul{0.13} / \ul{3.7} \\

desk-norm     &   0.46 / 3.9   &    0.47 /  4.0  &    \bd{0.09} / \bd{1.0}    &    0.61 / 5.2     &     3.66 / 31.0 & -- / --   &  -- / --  &    0.18 / 2.6  &  \ul{0.11} / \ul{1.1} \\
desk-fast       &   0.31 / 9.9   &    0.32 / 10.0    &    \bd{0.05} / \bd{1.9}    &    \ul{0.13} / \ul{4.2}    &    0.14 / 4.3    &    -- / --  &  -- / --   &  0.77 / 30.2  &  0.15 / 6.1 \\

sofa-norm     &   0.15 / 4.4 &    \ul{0.13} / \ul{3.8}      & \bd{0.06} / \bd{2.1} & 0.16 / 4.7 & 0.19 / 5.8 &  1.77 / 53.0 &-- / --  &  0.22  /  10.1  & 0.13 / 4.7 \\
sofa-fast       &   0.21 / 6.4 &    0.57  / 17.0    &     \bd{0.07} / \bd{2.2}   &    0.17 / 5.2     &    \ul{0.17} / \ul{5.0} & --  / --  &  -- /  -- & 0.60  / 22.2   &   0.38 / 14.4 \\

mount-norm   &   0.35 / 2.6   &    4.05 / 30.0   &   \bd{0.08} / \bd{0.7}     &    0.59 / 4.4    &    4.32 / 32.0   &  -- / --   &  -- / --   & \ul{0.09} / \ul{0.8}  &  \ul{0.09} / \ul{0.8} \\
mount-fast    &   2.11 / 5.2   &   -- / --   &     \bd{0.11} / \ul{3.7}   & 0.16 / 3.9 & \ul{0.13} / \bd{3.1} & -- / --  & -- / -- &  0.31 / 11.5 &  0.37 / 14.0 \\

hdr-normal       &   0.64  / 1.9  &   1.27 / 3.8  &   \bd{0.13} / \bd{0.5}     &    0.57 / \ul{1.7}     &    4.02 / 12.0    & -- / --  &-- / --  &  \ul{0.52} / 2.4 & 0.60 / 3.1 \\
hdr-fast         &   0.22 / 4.0   &   0.30 / 5.5  &   \bd{0.06} / \bd{1.2}  &  0.21 / 3.9  & \ul{0.20} / \ul{3.6} &-- / --  & -- / -- & 0.20 / 4.6 & 0.24 / 5.7 \\

\midrule
corr-dolly     &   1.03  / 80   &    1.88 / 146     &   \ul{0.56} / \ul{54}    &   1.13  / 88 & 1.58 / 123 &-- / --  & -- / -- & -- / -- & \bd{0.51} / \bd{53} \\
corr-walk      &   1.32 / 103   &    \ul{0.50} / \ul{39}     &   0.54 / 50 & \bd{0.43} / \bd{34} & 0.92 / 72 & -- / -- &-- / --  & -- / -- & 1.04 / 113 \\

school-dolly        &   0.73 / 92 &    1.42 / 179     &    \bd{0.11} / \bd{16} & 0.42 / 53 & 2.47  / 311 & 10.9 / 1371 &-- / --  & 4.61  / 651 & \ul{0.29} / \ul{41} \\
school-scooter      &   0.70 / 75   &    0.52 / \ul{56}     &    \bd{0.40} / \bd{47} & 0.59 / 63 & 1.30  / 139 & 9.21 / 983 & -- / -- & 0.55 / 66 & \ul{0.48} / 58 \\

units-dolly    &   7.64 / 1806    &    4.39 / 1039     &   \ul{1.52} / \ul{452}  & 3.43 / 812 & 5.84 / 1382 & -- / -- & -- / -- & -- / -- & \bd{0.48} / \bd{131} \\
units-scooter       &   6.22 /  1450   &    4.92 / 1147     &   \ul{1.67} / \ul{497} & 2.85 / 664 & 5.00 / 1166 & -- / -- &-- / --  & 3.69 /  1141 & \bd{0.88} / \bd{296} \\

\bottomrule
\end{tabular}
\caption{VECtor dataset~\cite{gao22vector} with MPE[\%/m] and ATE[cm]. Baseline numbers (except for DPVO and DPVO\largedagger) are taken from~\cite{chen2023esvio}.
The first eleven sequences are small-scale, while the last six sequences cover large-scale scenes. Since events of the large-scale sequences are sparsely distributed, these sequences benefit strongly from our patch selection mechanism (\cf~\cref{sec:ablation}).
Please note that methods utilizing frames have a considerable advantage due to the high-quality, global shutter FLIR Grasshopper3 camera with $1224\times 1024$ pixels, while the event camera is a Prophesee Gen3 with $640\times 480$ pixels from 2017. Moreover, the event stream exhibits significant artifacts due to the mounted infrared filter.}
\label{tab::VECTOR}
\end{table*}

\egroup
The VECtor dataset~\cite{gao22vector} provides stereo events and frames captured by two Prophesee Gen3 and two high-quality, global shutter FLIR Grasshopper3 cameras. It covers a variety of aggressive motion and illumination changes for both small and large-scale scenes.
As seen in~\cref{tab::VECTOR}, our proposed DEVO achieves outstanding results on most of the sequences under the fact that DEVO utilizes event data only while EVO~\cite{rebecq16EVO} and ESVO~\cite{zhou21ESVO} fail on 88\% and 76\% of the sequences, respectively. The baseline DPVO\largedagger~fails on four sequences and struggles with the fast and large-scale sequences.
Our method even beats ESVIO~\cite{chen2023esvio} on 70\% of the sequences, which utilize not only stereo events but also high-quality stereo frames and an IMU.
DEVO performs better on average than DPVO on large-scale sequences but is usually worse than DPVO on small-scale scenes. Please note that DPVO profits from the high-quality frames.

\pseudoparagraph{HKU}
\bgroup
\begin{table*}[tb]
\centering
\small

\footnotesize
\begin{tabular}{l aaa | aacc}
\toprule

Method & ORB3~\cite{campos21orb3} & VINS Fusion~\cite{qin19VINS} & DPVO~\cite{teed2022dpvo} & ESVIO~\cite{chen2023esvio} & PLEVIO~\cite{guan22plevio} & DPVO\largedagger & \bd{DEVO (ours)} \\

Modality    & Stereo VIO & Stereo VIO & Mono VO & Stereo EVIO & Mono EVIO & Mono EO\largedagger  & Mono EO  \\

\cmidrule(){2-8}
& MPE / ATE & MPE / ATE & MPE / ATE & MPE / ATE & MPE / ATE & MPE / ATE & MPE / ATE \\

\midrule

agg\_tran         &   0.15 / 9.5  &   0.11  / 6.9   &  0.07 / 9.66 &   0.10 / 6.3  & \ul{0.07} / \ul{4.8} & 0.12 / 15.42 &   \bd{0.06} / \bd{4.03} \\
agg\_rot          &   0.35 / 23.0   &   1.34 / 8.8    &  \bd{0.04} / \ul{3.70} & 0.17 / 11.0  & 0.23 / 15.0 & 0.28 / 22.02 & \ul{0.05} / \bd{3.58}  \\

agg\_flip\textsuperscript{*}         &   \bd{0.36} / \bd{14.0}   &   1.16 / 45.0     & 0.99 / 56.52  & \bd{0.36} / \bd{14.0}  & \ul{0.39} / \ul{15.0} & 1.16 / 60.26  &  0.71 / 44.90  \\
agg\_walk         &     -- / --     &     -- / --      & 1.17 / 107.4 & \bd{0.31} / \bd{27.0}  & \ul{0.42} / \ul{37.0}  & -- / -- &   0.90 / 88.27\myddag  \\

hdr\_circle\textsuperscript{*}       &   0.17 / 8.3   &   5.03 / 252.0   & 0.31 / 24.10 & \ul{0.16} / \ul{8.1} & \bd{0.14}  / \bd{6.8} & 1.19 / 67.60  & 0.39 / 21.30  \\
hdr\_slow\textsuperscript{*}          &   0.16 / 8.6  &   0.13 / 7.3      & 0.23 / 16.63 & \ul{0.11} / \ul{5.9}  & 0.13 / 6.9 & 0.36 / 24.01 &  \bd{0.08} / \bd{4.95}  \\

hdr\_tran\_rota   &   0.30 / 20.0   &   0.11 / 7.5     & 0.67 / 53.30 & \ul{0.10} / \ul{6.5} & \ul{0.10}  / 6.8  & 0.21 / 21.60 &  \bd{0.08} / \bd{5.91}   \\
hdr\_agg          &   0.29 / 28.0   &   1.21 / 118.0    & 0.29 / 36.10 & \bd{0.10}  / \bd{10.0} & \ul{0.14} / \ul{14.0} & 0.54 / 59.80 &  0.26 / 33.98  \\
hdr\_dark\_norm   &     -- / --     &   0.86 / 80.0     & -- / --  & \ul{0.42} / \ul{39.0} & 1.35 / 125.0 & 0.49 / 47.70 &   \bd{0.06} / \bd{6.19}  \\

\bottomrule
\end{tabular}
\caption{HKU dataset~\cite{chen2023esvio} with MPE[\%/m] and ATE[cm]. Baseline numbers (except for DPVO and DPVO\largedagger) are taken from~\cite{chen2023esvio}.
Note that the event-only VO methods EVO~\cite{rebecq16EVO} and ESVO~\cite{zhou21ESVO} fail on all sequences of the HKU dataset (\cf~\cite{chen2023esvio}).
Sequences marked with (*) contain 2-6 subsequent voxel grids with sensor detection failures on at least 80\% of the sensor array, making tracking hard at fast motion without additional modality. Results marked with (\textdaggerdbl) use top-$P$ sampling, otherwise DEVO fails with multinomial sampling.}
\label{tab::HKU}
\vspace{-12pt}
\end{table*}

\egroup
The HKU dataset~\cite{chen2023esvio} is a collection of stereo event data and frames from two DAVIS346~\cite{brandliDAVIS}, which contains scenes with extremely fast 6-dof motion and strong HDR.
In~\cref{tab::HKU}, we compare to other event- and frame-based VO and VIO. Our method outperforms all previous work on five out of nine sequences. Other event-only VO methods, EVO and ESVO, fail on all sequences.
DEVO fails with multinomial sampling on the very challenging sequence \emph{agg\_walk}, where the baseline DPVO\largedagger~and the two stereo VIO systems ORB-SLAM3 and VINS Fusion fail as well.
However, with top-$P$ sampling, DEVO can track the poses of \emph{agg\_walk} without failure.
Moreover, we beat DPVO, especially on HDR scenes. DPVO fails on one sequence. The results on this dataset demonstrate the exciting potential of event-only methods for challenging capturing conditions.

\pseudoparagraph{EDS} The EDS dataset~\cite{hidalgo22EDS} provides events from a handheld Prophesee Gen3.1. To the best of our knowledge, we provide the first results for pose estimation on EDS.
In~\cref{tab::EDS}, we compare our method with ORB-SLAM3, DPVO\largedagger, and DPVO. While we beat DPVO\largedagger~and perform similarly to DPVO on most sequences, we perform worse on \emph{peanuts\_dark}, \emph{peanuts\_light}, and \emph{peanuts\_run}. Our method performs clearly better than DPVO on \emph{rocket\_dark}, \emph{ziggy\_hdr}, and \emph{all\_chars} due to HDR and very fast motion.

\bgroup
\begin{table*}[tb]
\centering
\footnotesize

\begin{tabular}{l aaa aaa | ccc ccc}
\toprule

Method & \multicolumn{3}{c}{ORB-SLAM3~\cite{campos21orb3}}  & \multicolumn{3}{c|}{DPVO~\cite{teed2022dpvo}}  & \multicolumn{3}{c}{DPVO\largedagger}  &  \multicolumn{3}{c}{\bd{DEVO (ours)}} \\

Modality    & \multicolumn{3}{c}{Mono VO} & \multicolumn{3}{c|}{Mono VO} &  \multicolumn{3}{c}{Mono EO\largedagger}  & \multicolumn{3}{c}{Mono EO}  \\

\cmidrule(r){2-4} \cmidrule(lr){5-7} \cmidrule(lr){8-10} \cmidrule(lr){11-13}

& ATE & $\mathbf{R}_\text{rmse}$ & MPE   &  ATE & $\mathbf{R}_\text{rmse}$ & MPE   &  ATE & $\mathbf{R}_\text{rmse}$ & MPE    &  ATE & $\mathbf{R}_\text{rmse}$ & MPE     \\

\midrule

peanuts\_dark      &   6.15   &   11.40   &   0.49       &  \bd{1.26}  &  \bd{1.83} & \bd{0.12} &     5.76   &  8.55   &  0.52    &  \ul{4.78}   &  \ul{2.49}   &  \ul{0.30}     \\
peanuts\_light     &   27.26  &   6.88   &   1.14       & \bd{12.99}  &  \bd{2.66}  & \bd{0.44} &     69.77   &  13.83   &  2.36     &  \ul{21.07}   &  \ul{3.84}   &  \ul{0.75}     \\
peanuts\_run       &    \bd{16.83}   &    \bd{5.78}   &   \bd{0.19}       &  \ul{25.48}  &  \ul{11.19}  &  \ul{0.29} &     43.49   &  19.72   &  0.52  &   38.10   &  18.28   &  0.43     \\

rocket\_dark       &   \ul{10.12}   &   9.75   &   \ul{0.37}       &  27.41 & \ul{5.23} & 1.07 &     80.89   &  24.43 &  3.65    &  \bd{8.78} & \bd{4.16} 	& \bd{0.32}    \\
rocket\_light      &   \bd{32.53}   &   11.39   &   \bd{1.79}       & 63.11 & \ul{10.44} & 3.64 &     97.62   &  24.86   &  5.08    &  \ul{59.83} & \bd{9.28} &	\ul{3.40}    \\

ziggy             &   26.92   &   4.42   &   0.42       & \ul{14.86}  & \ul{3.45}  & \ul{0.22} &     23.79   &  6.29   &  0.36  &  \bd{11.84}   &  \bd{2.32}   &  \bd{0.15}     \\
ziggy\_hdr        &   81.98   &   17.67   &   1.13       &  66.17 & \ul{10.32} & 1.02 &     \ul{46.41}   &  15.84   &  \ul{0.72}  &  \bd{22.82}  &  \bd{9.07}   &  \bd{0.36}      \\
ziggy\_flying        &   20.57   &   8.02   &   1.33       &  \bd{10.85} &  \ul{3.66}  &  \ul{0.73}  &     34.51   &  9.04   &  2.05   &  \ul{10.92}   &  \bd{3.39}   &  \bd{0.71}     \\

all\_chars        &   \ul{21.37}   &   \ul{9.02}   &   \ul{0.27}       &  95.87 & 29.00 & 1.39 &     76.02   &  14.86   &  0.90  &   \bd{10.76}   &  \bd{3.62}   &  \bd{0.16}     \\

\bottomrule
\end{tabular}
\vspace{-2pt}
\caption{EDS dataset~\cite{hidalgo22EDS} with ATE[cm], $\mathbf{R}_\text{rmse}$[$\deg$], and mean position error MPE[\%/m]. Since events of the EDS dataset are more densely populated than in the other datasets, we reduce the time window for each event voxel grid by half. Our method performs clearly better than DPVO on \emph{rocket\_dark}, \emph{ziggy\_hdr}, and \emph{all\_chars} due to HDR and very fast motion.}
\label{tab::EDS}
\vspace{-2pt}
\end{table*}
\egroup

\pseudoparagraph{TUM-VIE} The TUM-VIE dataset~\cite{klenk21tumvie} provides stereo events from two HD Prophesee Gen4. We show the results in~\cref{tab::TUMVIE}.
We beat all event-only methods (DPVO\largedagger, EVO, and ESVO) and USLAM by large margins on all presented sequences. We even outperform DH-PTAM~\cite{soliman23DHTAM} on four out of five sequences, even though DH-PTAM utilizes all four cameras of the setup (stereo EVO). 
\bgroup

\begin{table*}[tb]

\centering
\small

\begin{tabular}{l l cccc c}
\toprule

Method                          & Modality   &  1d-trans &  3d-trans &  6dof &  desk & desk2 \\
\midrule

\rowcolor{Lightgray} 
ORB-SLAM3~\cite{campos21orb3}         & Stereo VIO      &    0.7   &     1.2  &   1.8  &  \bd{0.7}  &  2.5 \\
\rowcolor{Lightgray} 
BASALT~\cite{usenko2019Basalt}   & Stereo VIO      &    \bd{0.3}   &  \ul{0.9}  &   \ul{1.4}  &  1.6  &  1.1 \\
\rowcolor{Lightgray} 
DPVO~\cite{teed2022dpvo}   & Mono VO     &   \ul{0.5}   &  1.1   &  \bd{1.2}   &  \ul{1.2}   &  \bd{0.8} \\

\midrule

\rowcolor{Lightgray} 
DH-PTAM~\cite{soliman23DHTAM}    & Stereo EVO     &   10.3   &    \bd{0.7} &   2.4  &  1.6  & 1.5 \\
\rowcolor{Lightgray} 
USLAM~\cite{ultimateSlam18}      & Mono EVIO        &    3.9   &    4.7  &   35.3  &  19.5  & 34.1 \\
\rowcolor{Lightgray} 
ESVO~\cite{zhou21ESVO}          & Stereo EO       &    0.9   &    2.8  &   5.8  &  3.3  & 3.2 \\

EVO~\cite{rebecq16EVO}           & Mono EO         &    7.5   &    12.5  &   85.5  & 54.1  & 75.2 \\
DPVO\largedagger & Mono EO\largedagger & 2.3 & 8.2 & 7.9 & 5.1 & 3.7 \\
\bd{DEVO (ours)}                &  Mono EO   &   \ul{0.5}  & 1.1 & 1.6  & 1.7 & \ul{1.0}  \\

\bottomrule
\end{tabular}
\caption{TUM-VIE dataset~\cite{klenk21tumvie} with ATE[cm] on the \enquote{mocap} sequences. Baseline numbers (except for DPVO and DPVO\largedagger) are taken from~\cite{soliman23DHTAM}.
DEVO beats the event-only methods (DPVO\largedagger, EVO, and ESVO) by large margins on all sequences (at least 44\% lower ATE).}
\label{tab::TUMVIE}
\vspace{-8pt}
\end{table*}

\egroup

\pseudoparagraph{RPG} The RPG dataset~\cite{zhou18RPG} provides events from a handheld stereo DAVIS240C~\cite{brandliDAVIS}. \Cref{tab::RPG} shows that we outperform the event-only methods DPVO\largedagger, EVO, and ESVO (stereo EO) by large margins on all sequences. 
Moreover, we outperform USLAM (mono EVIO) and EDSO~\cite{hidalgo22EDS} (mono EVO) on all sequences, and DPVO on three of four sequences.
DEVO is only beaten on \emph{rpg\_bin} by DPVO (mono VO) and ORB-SLAM2~\cite{mur2017orb2} (stereo VO).
\bgroup

\begin{table*}[p]

\centering
\small

\begin{tabular}{l l cccc cccc}
\toprule

& &
\multicolumn{2}{c}{rpg\_bin} &  
\multicolumn{2}{c}{rpg\_boxes2} & 
\multicolumn{2}{c}{rpg\_desk2} &
\multicolumn{2}{c}{rpg\_monitor2} \\
\cmidrule(r){3-4} \cmidrule(lr){5-6} \cmidrule(lr){7-8} \cmidrule(l){9-10}

Method & Modality &
ATE & $\mathbf{R}_\text{rmse}$ & ATE & $\mathbf{R}_\text{rmse}$ & ATE & $\mathbf{R}_\text{rmse}$ & ATE & $\mathbf{R}_\text{rmse}$\\

\midrule

\rowcolor{Lightgray} 
ORB-SLAM2~\cite{mur2017orb2}     & Stereo VO     & \bd{0.7} & \ul{0.58}    & \ul{1.6} & 4.26     & 1.8 & 2.81     & \ul{0.8} & 3.70  \\ 
\rowcolor{Lightgray} 
ORB-SLAM2~\cite{mur2017orb2} & Mono VO       & 2.4 & 0.84    & 3.9 & 2.39     & 3.8 & 2.52     & 3.1 & 1.77  \\
\rowcolor{Lightgray} 
DSO~\cite{engel2017direct}  & Mono VO       & 1.1 & 2.12    & 2.0 & 2.14     & 10.0 & 63.5    & 0.9 & \bd{0.33}  \\
\rowcolor{Lightgray} 
DPVO~\cite{teed2022dpvo} & Mono VO &   \bd{0.7}   &  \bd{0.4}  &  1.62 &  \ul{1.78}   & 3.1 & \ul{1.04}   & 2.12 & 2.08  \\

\midrule

\rowcolor{Lightgray} 
USLAM~\cite{ultimateSlam18} & Mono EVIO     & 7.7 & 7.18    & 9.5 & 8.84     & 9.8 & 32.46    & 6.5 & 7.01  \\
\rowcolor{Lightgray} 
EDSO~\cite{hidalgo22EDS}    & Mono EVO      & 1.1 & 0.99    & 2.1 & 1.83     & \ul{1.5} & 1.87     & 1.0  & \ul{0.60}  \\
\rowcolor{Lightgray} 
ESVO~\cite{zhou21ESVO}      & Stereo EO     & 2.8 & 7.61    & 5.8 & 9.46     & 3.2 & 7.25     & 3.3 & 2.74  \\

EVO~\cite{rebecq16EVO}      & Mono EO &     13.2\textsuperscript{*}  & 50.26\textsuperscript{*}   & 14.2\textsuperscript{*} & 170.36\textsuperscript{*}  & 5.2 & 8.25     & 7.8 & 7.77    \\
DPVO\largedagger & Mono EO\largedagger & 4.00 & 3.23 & 4.20 & 2.87 & 3.05 & 1.45 & 2.35 & 2.79 \\

\bd{DEVO (ours)}    & Mono  EO      & \ul{1.03} & 0.86  & \bd{0.92} & \bd{0.70} & \bd{1.21} & \bd{0.95} &  \bd{0.71} &	1.04  \\

\bottomrule
\end{tabular}
\caption{RPG dataset~\cite{zhou18RPG} with ATE[cm] and $\mathbf{R}_\text{rmse}$[$\deg$]. Baseline numbers (except for DPVO and DPVO\largedagger) are taken from~\cite{huang23eVOSurvey}.
Results marked with (*) indicate failure after completing at most 30\% of the sequence.
DEVO outperforms the event-only methods (DPVO\largedagger, EVO, and ESVO) by large margins on all sequences (at least 63\% lower ATE). In addition, we attain an average ATE 88\% lower than USLAM (mono EVIO) and 28\% lower than EDSO (mono EVO).}
\label{tab::RPG}
\end{table*}

\egroup

\pseudoparagraph{MVSEC}
\begin{table*}[p]
\centering
\small

\begin{tabular}{ll cccc}
\toprule

Method & Modality & Indoor Fly1 & Indoor Fly2 & Indoor Fly3 & Indoor Fly4 \\

\cmidrule(){3-6}
& & MPE / ATE & MPE / ATE & MPE / ATE & MPE / ATE \\

\midrule

\rowcolor{Lightgray} 
ORB-SLAM3~\cite{campos21orb3} & Stereo VIO & 5.31 / 142.0 & 5.65 / 170.0  & 2.90 / 154.0 & 6.99 / 58.0 \\
\rowcolor{Lightgray} 
VINS Fusion~\cite{qin19VINS} & Stereo VIO & 1.50 / 40.0 & 6.98 / 210.0 & 0.73 / 39.0 & 3.62 / 30.0 \\
\rowcolor{Lightgray} 
DPVO~\cite{teed2022dpvo} & Mono VO & \textbf{0.16} / \textbf{4.8} & \textbf{0.15} / \textbf{6.3} & \textbf{0.08} / \textbf{4.6} & \textbf{0.30} / \textbf{3.2} \\

\midrule

\rowcolor{Lightgray} 
ESVIO~\cite{chen2023esvio} & Stereo EVIO & 0.94 / 25.0 & 1.00 / 30.0 & 0.47 / 25.0 & 5.55 / 46.0 \\
\rowcolor{Lightgray} 
USLAM~\cite{ultimateSlam18} & Mono EVIO & -- / -- & -- / --  & -- / -- & 2.77 / 23.0 \\
\rowcolor{Lightgray} 
PLEVIO~\cite{guan22plevio} & Mono EVIO & 1.35 / 36.0 & 1.00 / 30.0 & 0.64 / 34.0 & 5.31 / 44.0 \\
\rowcolor{Lightgray} 
ESVO~\cite{zhou21ESVO} & Stereo EO & 4.00 / 107.0 & 3.66 / 110.0 & 1.71 / 91.0 & -- / -- \\

EVO~\cite{rebecq16EVO} & Mono EO & 5.09 / 136.0 & -- / -- & 2.58 / 137.0 & -- / -- \\
DPVO\largedagger & Mono EO\largedagger & -- / -- & -- / -- & -- / -- & -- / -- \\

\textbf{DEVO (ours)} & Mono EO & \underline{0.26} / \underline{7.76} & \underline{0.32} / \underline{13.30} & \underline{0.19} / \underline{10.72} & \underline{1.08} / \underline{12.57} \\
\bottomrule

\end{tabular}
\caption{MVSEC dataset~\cite{zhu18mvsec} with MPE[\%/m] and ATE[cm]. Baseline numbers (except for DPVO and DPVO\largedagger) are taken from~\cite{chen2023esvio}.
DEVO is the first event-only method that does not fail on any sequence of MVSEC. DEVO performs substantially better than all prior work, except DPVO, on all sequences. DEVO performs worse than DPVO, which might be due to the biased polarity ratio of the event data.}
\label{tab::MVSEC}
\end{table*}
The MVSEC dataset~\cite{zhu18mvsec} provides stereo events and frames from two DAVIS346.
We compare to previous work~\cite{chen2023esvio} on the indoor flying room.
In~\cref{tab::MVSEC}, we surpass all related work, except DPVO, by large margins on all sequences. The event-only methods EVO and ESVO fail on at least one sequence.
We attain an average ATE of 65\% lower than ESVIO, which utilizes stereo events, stereo frames, and IMU.
DEVO performs worse than frame-based DPVO, which might be due to the noisy event data of MVSEC. The event stream of MVSEC exhibits a negative-to-positive polarity ratio of 3.2, which is by far the largest among all seven real-world datasets: EDS (1.5), FPV (1.5), VECtor (1.2), HKU (0.8), RPG (0.7), TUM-VIE (0.7). This noisy event stream also leads to artifacts in the E2VID reconstructions, resulting in failures of DPVO\largedagger~on all sequences.

\subsection{Ablation Study}\label{sec:ablation}
We perform ablation experiments on seven real-world event VO benchmarks to study our contributions. We run each sequence five times, take the median per sequence, and report the average of medians per dataset.

\pseudoparagraph{Photometric voxel augmentation}
In \cref{tab::ablation_randaug}, DEVO benefits from photometric augmentation on all datasets, except on VECtor~\cite{gao22vector}, where it shows a less than 3.6\% accuracy drop.
Voxel grids of the training dataset follow an ideal event generation model (\cf \cref{eq:evModel}). One prominent effect of our photometric augmentation is to reduce event density. Particularly, the UZH-FPV~\cite{delmer19FPV} dataset benefits strongly.
\bgroup

\begin{table}[H]
\centering
\footnotesize

\vspace{5pt}
\begin{tabular}{l cc cc}
\toprule
Dataset & \multicolumn{2}{c}{w/o photo. Aug.} & \multicolumn{2}{c}{\textbf{with photo. Aug.}} \\

\cmidrule(r){2-3} \cmidrule(l){4-5}
& $\overline{\text{ATE}}$ & \#fails  & $\overline{\text{ATE}}$ & \#fails  \\
\midrule

UZH-FPV~\cite{delmer19FPV}      & 132.18 & 1        & \bd{107.01} & 2    \\
VECtor~\cite{gao22vector}       & \bd{42.52} & 0    & 44.03 & 0     \\

HKU~\cite{chen2023esvio}        & 17.26 & 7         & \bd{15.60} & 6     \\

EDS~\cite{hidalgo22EDS}         & 23.24 & 4         & \bd{20.99} & 4  \\
TUM-VIE~\cite{klenk21tumvie}    & 1.38 & 0          & \bd{1.20} & 0 \\

RPG~\cite{zhou18RPG}            & 1.00 & 0          & \bd{0.97} & 0      \\
MVSEC~\cite{zhu18mvsec}         & 13.37 & 1         & \bd{11.08} & 0      \\

\midrule
All datasets                    & 32.99 & 13           & \bd{28.70} & \bd{12} \\

\bottomrule

\end{tabular}
\caption{Ablation study of photometric voxel augmentation. Average $\overline{\text{ATE}}$[cm] and total number of failures (\#fails) for seven real-world datasets. Our proposed photometric augmentation improves accuracy and robustness.}
\label{tab::ablation_randaug}
\vspace{-8pt}
\end{table}
\egroup

\pseudoparagraph{Patch selection and sampling}
\bgroup

\begin{table*}[p]

\centering
\footnotesize
\begin{tabular}{l cc | c | cccc}
\toprule

Training  & random & gradient & \multicolumn{5}{c}{\bd{score map (ours)}} \\
\cmidrule(r){2-2}\cmidrule(lr){3-3}\cmidrule(l){4-8}

Inference & random & pool-grid & $3P$-random & top-$P$ samp. & \multicolumn{3}{c}{\bd{multinomial sampling}} \\
\cmidrule(r){2-2}\cmidrule(lr){3-3}
\cmidrule(lr){4-4}\cmidrule(lr){5-5}\cmidrule(l){6-8}

Settings & & & & pool-grid & w/o pool. & w/o grid & \bd{pool-grid} \\
\midrule

& $\overline{\text{ATE}}$ / \# & $\overline{\text{ATE}}$ / \# & $\overline{\text{ATE}}$ / \# & $\overline{\text{ATE}}$ / \# & $\overline{\text{ATE}}$ / \# & $\overline{\text{ATE}}$ / \# & $\overline{\text{ATE}}$ / \# \\
\midrule

UZH-FPV~\cite{delmer19FPV}      & 190.11 / 2 & 128.93 / 0       & 121.19 / 4 & 111.85 / 0       & 142.63 / 3 & \ul{109.33} / 2  &  \bd{107.01} / 2 \\

VECtor~\cite{gao22vector}       & 70.16 / 0 & 106.27 / 0         & 63.97 / 0 & 68.01 / 0         & 64.40 / 0   & \ul{52.92} / 0   & \bd{44.03} / 0 \\
HKU~\cite{chen2023esvio}        &  21.56 / 10 & \ul{15.67} / 8        & 18.62 / 8  & 18.99 / 2        & 20.87 / 7 & 18.18 / 5  & \bd{15.60} / 6   \\

EDS~\cite{hidalgo22EDS}         & \ul{24.08} / 4 & 31.64 / 5         & 27.95 / 3 & 33.63 / 8         & 36.44 / 7 & 28.13 / 4 & \bd{20.99} / 4  \\

TUM-VIE~\cite{klenk21tumvie}    & 1.49 / 0 & \bd{1.16} / 0      & 1.18 / 0 & 1.25 / 0           & 1.18 / 0 & \ul{1.17} / 0 & 1.20 / 0  \\

RPG~\cite{zhou18RPG}            & 1.02 / 0   & \bd{0.92}	/ 0         & 1.14 / 0   & 1.04 / 0         & 1.18 / 0  & 1.11 / 0  & \ul{0.97} / 0 \\
MVSEC~\cite{zhu18mvsec}         & \bd{11.02} / 0 & 13.01 / 2         & 27.10 / 2  &  13.11 / 4        & 13.61 / 1   & 34.16 / 5   & \ul{11.08} / 0 \\

\midrule
All datasets                   & 45.63 / 16 & 42.51 / 15         & 37.31 / 17 & 35.41 / \ul{14}      & 40.04 / 18 & \ul{35.00} / 16 & \bd{28.70} / \bd{12} \\
\bottomrule

\end{tabular}
\caption{Ablation study on patch selection methods during training and respective patch sampling at inference evaluated on seven real-world datasets. Average $\overline{\text{ATE}}$[cm] and total number of failures (\#) for each dataset.
DPVO~\cite{teed2022dpvo} employs random patch selection during training and inference for dense RGB data, which does not work well for events. Our proposed grid-based pooled multinomial sampling on the learned score map performs best on four of seven datasets with the smallest number of failures.
Pool refers to $4\times 4$ average pooling before sampling. Refer to~\cref{sec:method} for a detailed description of all sampling methods.}
\label{tab::ablation_patch_selector}
\end{table*}

\egroup
\Cref{tab::ablation_patch_selector} shows that our learned score map improves the pose tracking accuracy over random or gradient-based sampling during training and inference.
It also shows that our grid-based pooled multinomial sampling on the learned score map achieves the best results for the majority of the datasets with the smallest number of failures.
Our multinomial sampling without grid ranks second on average ATE, but it is less robust, especially on MVSEC.
While the gradient-based method performs slightly better on RPG and TUM-VIE, our proposed method shows the best accuracy on more challenging scenes.  
Top-$P$ sampling performs significantly worse than our method. Employing the grid improves robustness (12 failures w/ grid \vs 16 failures w/o), which is expected for sparse event data. The proposed average pooling before sampling increases accuracy and robustness on average.
\vspace{-1pt}
\section{Conclusion}
\vspace{-2pt}
We present DEVO, a monocular event-only VO with strong performance on seven real-world benchmarks. Our method tracks sparse event patches over time, extracted by our learned patch selector and pooled multinomial sampling mechanism. 
Our results clearly show the benefit of using event cameras for VO in challenging scenarios. We show that learning-based VO for events is a promising direction. Our work is the first to train on a huge, simulated training dataset of events showing strong generalization to the real world. Future work could investigate using even larger, more diverse datasets for training or more realistic noise models for event simulation. Another interesting direction is to combine DEVO with other sensors.
\vfill
\pseudoparagraph{Acknowledgments}
This work was supported by the ERC Advanced Grant SIMULACRON.

\clearpage
{
    \small
    \bibliographystyle{ieeenat_fullname}
    \bibliography{main}
}

\end{document}